\documentclass[sigconf]{acmart}

\usepackage{booktabs}
\usepackage{graphicx}
\usepackage[tight,footnotesize]{subfigure}
\usepackage{soul}
\usepackage[utf8]{inputenc}
\usepackage{multirow,makecell,footmisc,url}
\usepackage{amsmath,amssymb,amsthm,bm}
\usepackage[ruled]{algorithm2e}
\usepackage{latexsym}
\usepackage{marvosym}
\usepackage{hyperref}

\copyrightyear{2018} 
\acmYear{2018} 
\setcopyright{acmcopyright}
\acmConference[MM '18]{2018 ACM Multimedia Conference}{October 22--26, 2018}{Seoul, Republic of Korea}
\acmBooktitle{2018 ACM Multimedia Conference (MM '18), October 22--26, 2018, Seoul, Republic of Korea}
\acmPrice{15.00}
\acmDOI{10.1145/3240508.3240543}
\acmISBN{978-1-4503-5665-7/18/10}

\begin{document}

\title{Deep Triplet Quantization}

\author{Bin Liu$^{1}$, Yue Cao$^{1}$, Mingsheng Long$^{1}$(\Letter), Jianmin Wang$^{1}$, and Jingdong Wang$^{2}$}
\affiliation{%
\institution{
    $^{1}$School of Software, Tsinghua University, Beijing 100084, China \\
    $^{1}$Beijing National Research Center for Information Science and Technology \\
    $^{2}$Microsoft Research Asia}
}
\email{{liubinthss,caoyue10}@gmail.com, {mingsheng,jimwang}@tsinghua.edu.cn, jingdw@microsoft.com}

\renewcommand{\shortauthors}{Bin Liu et al.}
\graphicspath{{figures/}}

\begin{abstract}
Deep hashing establishes efficient and effective image retrieval by end-to-end learning of deep representations and hash codes from similarity data. We present a compact coding solution, focusing on deep learning to quantization approach that has shown superior performance over hashing solutions for similarity retrieval. We propose Deep Triplet Quantization (DTQ), a novel approach to learning deep quantization models from the similarity triplets. To enable more effective triplet training, we design a new triplet selection approach, Group Hard, that randomly selects hard triplets in each image group. To generate compact binary codes, we further apply a triplet quantization with weak orthogonality during triplet training. The quantization loss reduces the codebook redundancy and enhances the quantizability of deep representations through back-propagation. Extensive experiments demonstrate that DTQ can generate high-quality and compact binary codes, which yields state-of-the-art image retrieval performance on three benchmark datasets, NUS-WIDE, CIFAR-10, and MS-COCO.
\end{abstract}



\maketitle

\section{Introduction}

Approximate nearest neighbors (ANN) search has been widely applied to retrieve large-scale multimedia data in search engines and social networks.
Due to the low storage cost and fast retrieval speed, learning to hash has been increasingly popular in the ANN research community, which transforms high-dimensional media data into compact binary codes and generates similar binary codes for similar data items. 
This paper will focus on data-dependent hashing schemes for efficient image retrieval, which have achieved better performance than data-independent hashing methods, e.g. Locality-Sensitive Hashing (LSH) \cite{cite:VLDB99LSH}.

A rich line of hashing methods have been proposed to enable efficient ANN search using Hamming distance \cite{cite:NIPS09BRE,cite:CVPR11ITQ,cite:ICML11MLH,cite:CVPR12MIH,cite:CVPR12KSH,cite:TPAMI12SSH,cite:SIGIR14LFH}.
Recently, deep hashing methods \cite{cite:AAAI14CNNH,cite:CVPR15DNNH,cite:CVPR15SDH,cite:CVPR15DH,cite:AAAI16DHN,cite:IJCAI16DPSH,cite:CVPR2016DSH,cite:ECCV16BDNN,cite:ICCV17HashNet,cite:ICCV17SSBC} have shown that both image representation and hash coding can be learned more effectively using deep neural networks, resulting in state-of-the-art results on many benchmark datasets. 
In particular, it proves crucial to jointly preserve similarity and control quantization error of converting continuous representations to binary codes \cite{cite:AAAI16DHN,cite:IJCAI16DPSH,cite:CVPR2016DSH,cite:ICCV17HashNet}. 
However, a pivotal weakness of these deep hashing methods is that they first learn continuous deep representations, and then convert them into hash codes by a separated binarization step.
By \emph{continuous relaxation}, i.e. solving the original discrete optimization of hash codes with continuous optimization, the optimization problem deviates significantly from the original hashing objective. As a result, these methods cannot learn \emph{exactly} compact binary hash codes in their optimization.

To address the limitation of continuous relaxation, Deep Quantization Network (DQN) \cite{cite:AAAI16DQN} and Deep Visual-Semantic Quantization (DVSQ) \cite{cite:CVPR17DVSQ} are proposed to integrate quantization method \cite{cite:TPAMI14OPQ,cite:ICML14CQ,cite:CVPR16SQ} and deep learning.
The quantization method represents each point by a short binary code formed by the index of the nearest center, which can generate natively binary codes and empirically achieve better performance than hashing methods for ANN search. 
However, previous deep quantization methods are either point-wise method that relies on expensive class-label information, or pairwise method that cannot capture the \emph{relative} similarity between images, i.e. a pair of images should not be seen as \emph{absolutely} similar or dissimilar. In other words, there should be a continuous spectrum from very similar to very dissimilar relations.

Recently, the \emph{triplet loss} \cite{cite:NIPS12HDML} has been studied for computer vision problems. The triplet loss captures the \emph{relative} similarity, which only brings anchor images closer to positive samples than to negative samples, hence it fits the ranking tasks naturally and achieves better performance than point-wise and pairwise losses for retrieval tasks.
However, how to enable effective triplet training for deep learning to quantization with only pairwise similarity available still remains a challenge.
Note that, without effective triplet selection, previous deep hashing method with triplet loss \cite{cite:CVPR15DNNH} cannot achieve superior results.
Hence, how to select good triplets for effective training in deep quantization also remains an open problem.

Towards these open problems, this paper presents Deep Triplet Quantization (DTQ) for efficient and effective image retrieval, which introduces a novel triplet training strategy to deep quantization, offering superior retrieval performance.
The proposed solution is comprised of four main components:
{1)} a novel triplet selection module, \emph{Group Hard}, to mine good triplets for effective triplet training;
{2)} a standard deep convolutional neural network (CNN), e.g. AlexNet or ResNet, for learning deep representations;
{3)} a well-specified triplet loss for pulling together similar pairs and pushing away dissimilar pairs;
and {4)} a novel triplet quantization loss with weak orthogonality constraint for converting the deep representations of different samples (such as the anchor, positive and negative samples) in the triplets into $B$-bit compact binary codes. The weak-orthogonality reduces the redundancy of codebooks and controls the quantizability of deep representations.
Comprehensive empirical evidence shows that the proposed DTQ can generate compact binary codes and yield state-of-the-art retrieval results on three image retrieval benchmarks, NUS-WIDE, CIFAR-10, and MS-COCO.

\section{Related Work}

Existing hashing methods can be categorized into unsupervised hashing and supervised hashing \cite{cite:NIPS09BRE,cite:CVPR11ITQ,cite:ICML11MLH,cite:CVPR12MIH,cite:CVPR12KSH,cite:TPAMI12SSH,cite:CVPR13HBS,cite:CVPR13BP,cite:ICML14CBE,cite:SIGIR14LFH,cite:TKDE15OCKmeans}. Please refer to \cite{cite:TPAMI2018HashSurvey} for a comprehensive survey.

Unsupervised hashing methods learn hash functions to encode data points to binary codes by training from unlabeled data. Typical learning criteria include reconstruction error minimization \cite{cite:AI07SemanticHashing,cite:CVPR11ITQ,cite:TPAMI11PQ} and graph learning \cite{cite:NIPS09SH,cite:ICML11AGH}.
Supervised hashing explores supervised information (e.g. given similarity or relevance feedback) to learn compact hash codes. Binary Reconstruction Embedding (BRE) \cite{cite:NIPS09BRE} pursues hash functions by minimizing the squared errors between the distances of data points and the distances of their corresponding hash codes. Minimal Loss Hashing (MLH) \cite{cite:ICML11MLH} and Hamming Distance Metric Learning \cite{cite:NIPS12HDML} learn hash codes by minimizing the triplet loss functions based on similarity of data points. Supervised Hashing with Kernels (KSH) \cite{cite:CVPR12KSH} and Supervised Discrete Hashing (SDH) \cite{cite:CVPR15SDH} build discrete binary codes by minimizing the Hamming distances across similar pairs and maximizing the Hamming distances across dissimilar pairs.

As deep convolutional networks \cite{cite:NIPS12CNN,cite:CVPR16DRL} yield advantageous performance on many computer vision tasks, deep hashing methods have attracted wide attention recently. CNNH \cite{cite:AAAI14CNNH} adopts a two-stage strategy in which the first stage learns binary hash codes and the second stage learns a deep-network based hash function to fit the codes. DNNH \cite{cite:CVPR15DNNH} improved  CNNH with a simultaneous feature learning and hash coding  pipeline such that deep representations and hash codes are optimized by the triplet loss. DHN \cite{cite:AAAI16DHN} and HashNet \cite{cite:ICCV17HashNet} improve DNNH by jointly preserving the pairwise semantic similarity and controlling the quantization error by simultaneously optimizing the pairwise cross-entropy loss and quantization loss via a multi-task approach.

Quantization methods \cite{cite:AAAI16DQN,cite:CVPR17DVSQ} represent each point by a short code formed by the index of the nearest center, have been shown to give more powerful representation ability than hashing for approximate nearest neighbor search. 
To our best knowledge, Deep Quantization Network (DQN) \cite{cite:AAAI16DQN} and Deep Visual-Semantic Quantization (DVSQ) \cite{cite:CVPR17DVSQ} are the only two prior works on deep learning to quantization. 
DQN jointly learns deep representations via a pairwise cosine loss and a product quantization loss \cite{cite:TPAMI11PQ} for generating compact binary codes. 
DVSQ proposes a pointwise adaptive-margin Hinge loss exploring class labels, and a visual-semantic quantization loss for inner-product search.

There are several key differences between our work and previous deep learning to quantization methods. 
{1)} Our work introduces a novel triplet training strategy to deep quantization framework for efficient similarity retrieval. It is worth noting that DTQ can learn compact binary codes when only the \emph{relative} similarity information is available, which is more general than the label-based quantization method DVSQ.
{2)} During the triplet learning procedure, DTQ proposes a novel triplet mining strategy, \emph{Group Hard}, resulting in faster convergence and better search accuracy.
{3)} DTQ proposes a novel triplet quantization loss with weak orthogonality constraint to reduce coding redundancy. An end-to-end architecture to join the above three terms yield both efficient and effective image retrieval.

\begin{figure*}[htbp]
  \centering
  \includegraphics[width=0.99\textwidth]{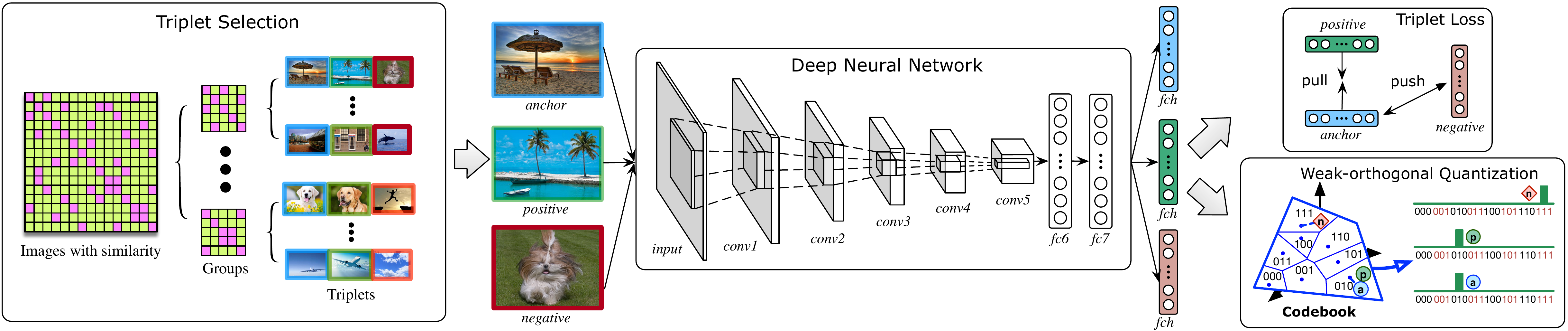}
  \caption{The proposed Deep Triplet Quantization (\textbf{DTQ}) model consists of four main components:
\textbf{1)} a novel triplet selection module, \emph{Group Hard}, to mine good triplets for effective triplet training and faster convergence;
\textbf{2)} a standard deep convolutional neural network (CNN), e.g. AlexNet, VGG or ResNet, for learning deep representations;
\textbf{3)} a well-specified triplet loss for pulling together similar pairs and pushing away dissimilar pairs;
and \textbf{4)} a novel triplet quantization loss with weak orthogonality constraint for converting the deep representations of different samples (the anchor, positive and negative samples) in the triplets into $B$-bit compact binary codes and controlling the quantizability of the deep representations.
\textit{Best viewed in color}.}
  \label{fig:DTQ}
\end{figure*}

\section{Deep Triplet Quantization}

In similarity retrieval, we are given $N$ training points $\mathcal{X} = \{{\bm x}_i\}_{i=1}^N$, where some pairs of points ${\bm x}_i$ and ${\bm x}_j$ are given with \emph{pairwise} similarity labels $s_{ij}$, where $s_{ij} = 1$ if ${\bm x}_i$ and ${\bm x}_j$ are similar while $s_{ij} = 0$ if ${\bm x}_i$ and ${\bm x}_j$ are dissimilar.
The goal of deep learning to quantization is to learn a composite quantizer $q:{\bm{x}} \mapsto {\bm{b}} \in {\left\{ { 0,1} \right\}^B}$ from input space to binary coding space $\{0,1\}^B$ through deep networks, which encodes each point ${\bm x}$ into $B$-bit binary code ${\bm b} = q({\bm x})$ such that the supervision in the training data can be maximally preserved. In supervised hashing, the similarity pairs $\{({\bm x}_i, {\bm x}_j, s_{ij}): s_{ij} \in \mathcal{S}\}$ are readily available from semantic labels or relevance feedbacks from click-through data in many image search engines.

We propose Deep Triplet Quantization (\textbf{DTQ}), an end-to-end architecture to join deep learning and quantization, as shown in Figure \ref{fig:DTQ}. DTQ has four key components:
1) a novel triplet selection module, \emph{Group Hard}, to mine a appropriate number of good triplets for effective triplet training;
2) a standard deep convolutional neural network (CNN), e.g. AlexNet, VGG, or ResNet, for learning deep representations;
3) a well-specified triplet loss for pulling together similar pairs and pushing away dissimilar pairs;
and 4) a novel triplet quantization loss with weak orthogonality constraint for converting deep representations of different samples (the \emph{anchor}, \emph{positive} and \emph{negative} samples) in triplets into $B$-bit compact binary codes and controlling the quantizability of the deep representations.

\subsection{Triplet Training}

We train a convolutional network from image triplets $\mathcal{T}$=$\{{\bm t}_i\}^{N_t}_{i=1}$. Each triplet ${\bm t}_i$=$\langle{\bm x}^a_i, {\bm x}^p_i, {\bm x}^n_i\rangle$ is constructed from pairwise similarity data $\{({\bm x}_i, {\bm x}_j, s_{ij}): s_{ij} \in \mathcal{S}\}$ as follows: for each \emph{anchor} image ${\bm x}_i^a$, we find a \emph{positive} image ${\bm x}_i^p$ with $s_{ap} = 1$ (${\bm x}_i^a$ and ${\bm x}_i^p$ are similar), and a \emph{negative} image ${\bm x}_i^n$ with $s_{an} = 0$ (${\bm x}_i^a$ and ${\bm x}_i^n$ are dissimilar).
Given a triplet ${\bm t}_i$=$\langle{\bm x}^a_i, {\bm x}^p_i, {\bm x}^n_i\rangle$, the deep network maps the triplet ${\bm t}_i$ into a learned feature space with $f({\bm t}_i)$=$\langle{\bm z}^a_i, {\bm z}^p_i, {\bm z}^n_i\rangle$.
We ensure that an anchor image ${\bm x}^a_i$ is closer to all positive images ${\bm x}^p_i$ than to all negative images ${\bm x}^n_i$.
And the \emph{relative} similarity between the images in triplets, ${\bm x}^a_i, {\bm x}^p_i, {\bm x}^n_i$, are measured by the Euclidean distances between their deep features, ${\bm z}^a_i, {\bm z}^p_i$, ${\bm z}^n_i$.
Thus the triplet loss is
\begin{equation}\label{eqn:L}
	L = \sum\limits_{i = 1}^{{N_t}} L_i = \sum\limits_{i = 1}^{{N_t}} {\max \left( {0,\delta  - \left\| {{\bm{z}}_i^a - {\bm{z}}_i^n} \right\|_2^2 + \left\| {{{\bm{z}}_i^a} - {\bm{z}}_i^p} \right\|_2^2} \right)} ,
\end{equation}
where $\delta$ is a margin that is enforced between positive and negative pairs, and $\mathcal{T}$ is the set of cardinality $N_t$ for all possible triplets in the training set. 
Compared to the widely-used pointwise and pairwise metric-learning losses \cite{cite:AAAI16DQN,cite:CVPR17DVSQ} in previous deep quantization methods, the triplet loss \eqref{eqn:L} only requires anchor samples to be more similar to positive samples than to negative samples, by a specifically margin. 
This establishes a \emph{relative} similarity relation between images, thus is much more reasonable than the \emph{absolute} similarity relation used in previous pointwise or pairwise approaches.

However, as the dataset gets larger, the number of triplets grows cubically, and generating all possible triplets would result in many easy triplets with $L_i = 0$ in Eq.~\eqref{eqn:L}, which would not contribute to the training and suffer from slower convergence.
Note that, without a sophisticated triplet selection procedure, previous deep hashing methods with the triplet loss \cite{cite:CVPR15DNNH} cannot achieve superior performance.
Consequently, it is crucial to mine good triplets for effective triplet training and faster convergence.
In this paper, we propose a novel triplet selection module, \emph{Group Hard}, to ensure the number of mined valid triplets is neither too big nor too small. The core idea is that we first randomly split the training data into several groups $\{G_i\}_{i=1}^{|G|}$, then \emph{randomly} select one \emph{hard} negative sample for each anchor-positive pair in one group. The proposed  triplet selection method is formulated as
\begin{equation}\label{eqn:select}
{\mathcal T} = \bigcup\limits_{i = 1}^{|G|} {\bigcup\limits_{a \in {G_i}} {\bigcup\limits_{p \in G_i^p} {{\rm{rand}}\left( {G_i^n} \right)} } },
\end{equation}
where $G_i^p={\left\{ {p \in {G_i}:p \ne a,{s_{ap}} = 1} \right\}}$ is the group of positive samples consisting of the samples similar to the anchor $a$ in the $i$th group, $\text{rand}( {G_i^n}$) is the random function that randomly chooses one negative sample from the group of hard negative samples $G_i^n$=${\left\{ {n \in {G_i}:\delta  - \left\| {{\bm{z}}_i^a - {\bm{z}}_i^n} \right\|_2^2 + \left\| {{{\bm{z}}_i^a} - {\bm{z}}_i^p} \right\|_2^2 > 0,{s_{an}} =  0} \right\}}$.
Here \emph{hard} negative sample ${\bm x}_i^n$ is defined as having non-zero loss value for a triplet ${\bm t}_i$=$\langle{\bm x}^a_i, {\bm x}^p_i, {\bm x}^n_i\rangle$. 
Note that, mining only the triplets with the hardest negative images would select the outliers in the dataset and make it unable to learn ground truth relative similarity. Thus in this paper, the proposed DTQ only selects the negative examples with moderate hardness, based on the random sampling $\text{rand}(G_i^n)$ instead of $\text{argmax}(G_i^n)$ in Eq.~\eqref{eqn:select}.

As the training proceeds, the average of triplet loss becomes smaller and the size of the hard triplets reduces. To ensure that there are enough hard triplets each epoch for effective triplet training, we design a decay strategy for the size of groups $|G|$ as: if the actual number of valid hard triplets is lower than the minimum number of the valid hard triplets (the constant MIN$\_$TRIPLETS in Algorithm~\ref{algorithm:DTQ}), the size of the groups is halved until $|G|=1$.

\textbf{Complexity:} Similar to previous work on triplet training \cite{zhao2017deeply}, we can prune the triplets with zero losses ($L_i = 0$), resulting a valid triplet set $\mathcal{T}$ whose size $|{\mathcal T}|$ is much smaller than the possible number $N^3$ of triplets.
Through the proposed \emph{Group Hard} selection strategy that chooses one negative sample for each anchor-positive pair in each group, the number of the candidate triplets for training is further reduced to $|\mathcal{T}|/\left| G \right|$. 
Furthermore, all the selected triplets are \emph{hard} triplets ($L_i>0$ in Eq.~\eqref{eqn:L}), and the total amount can be controlled in a suitable range by adjusting the number of groups $G$, resulting in effective triplet training and higher retrieval accuracy.

\subsection{Weak-Orthogonal Quantization}

While triplet training with Group Hard selection enables effective image retrieval, efficient image retrieval is enabled by a novel triplet quantization model.
As each batch used for training the deep neural networks is comprised of triplets, the proposed quantization model should be compatible with the triplet training.
For the $i$th triplet, each image representation ${\bm{ z}}^*_i$, where ${*}\in\{a,p,n\}$, is quantized using a set of $M$ codebooks ${\bm C}^* = [{\bm C}^*_1,\ldots,{\bm C}^*_M]$, where each codebook ${\bm C}^*_m$ contains $K$ codewords ${\bm C}^*_m = [{\bm C}^*_{m1},\ldots,{\bm C}^*_{mK}]$, and each codeword ${\bm C}^*_{mk}$ is a $D$-dimensional cluster-centroid vector as in K-means. Corresponding to the $M$ codebooks, we partition the binary codewords assignment vector ${\bm b}^*_i$ into $M$ $1$-of-$K$ indicator vectors ${\bm b}^*_i = [{\bm b}^*_{1i}; \ldots; {\bm b}^*_{Mi}]$, and each indicator vector ${\bm b}^*_{mi}$ indicates which one (and only one) of the $K$ codewords in the $m$th codebook is used to approximate the $i$th data point ${\bm z}_i^\ast$.
To enable knowledge sharing across the anchors, positive and negative samples in the triplets, we propose a \emph{triplet quantization} approach by sharing the codebooks $\{{\bm C}_m^* = {\bm C}_m\}_{m=1}^M$ across different samples in all triplets. To mitigate the degeneration issue of K-means, we further propose a \emph{weak orthogonality} penalty across the $M$ codebooks, which potentially reduces the redundancy of the multiple codebooks and improves the compactness of the binary codes.
The proposed triplet quantization model with weak-orthogonal constraint is defined as
\begin{equation}\label{eqn:Q}
	Q = \sum\limits_{i = 1}^{{N_t}} {\sum\limits_{* \in \{ a,p,n\} } {\left\| {{\bm{z}}_i^* - \sum\limits_{m = 1}^M {{{\bm{C}}_m}{\bm{b}}_{mi}^*} } \right\|_2^2} }  + \gamma \sum\limits_{m=1}^M \sum\limits_{m'=1}^M {\left\| {{\bm{C}}_m^{\sf T}{{\bm{C}}_{m'}} - {\bm{I}}} \right\|_F^2} 
\end{equation}
where ${\left\| {{{\bm{b}}_{{m}i}^{*}}} \right\|_0} = 1,{{\bm{b}}_{{m}i}^{*}} \in {\left\{ {0,1} \right\}^K}$, ${\left\|  \cdot  \right\|_0}$ is the $\ell_0$-norm that simply counts the number of the vector's nonzero elements, and $\gamma$ is the hyper-parameter that controls the degree of orthogonality. The $\ell_0$ constraint guarantees that only one codeword in each codebook can be activated to approximate the input data, which leads to compact binary codes. The underlying reason of using $M$ codebooks instead of single codebook to approximate each input data point is to further minimize the quantization error, while single codebook yields significantly lossy compression and large performance drop.

\subsection{Deep Triplet Quantization}

We enable efficient and effective image retrieval in an end-to-end architecture by integrating the triplet training procedure \eqref{eqn:L}, triplet selection module \eqref{eqn:select} and the weak-orthogonal quantization \eqref{eqn:Q} in a unified deep triplet quantization (DTQ) model as 
\begin{equation}\label{eqn:model}
	\mathop {\min }\limits_{{\Theta},{\bm{C}},{\bm{B}^*}} L + \lambda Q,
\end{equation}
where $\lambda > 0$ is a hyper-parameter between the triplet loss $L$ and the triplet quantization loss $Q$, and $\Theta$ denotes the set of learnable parameters of the deep network. Through joint optimization problem \eqref{eqn:model}, we can learn the binary codes by jointly preserving the similarity via triplet learning procedure and controlling the quantization error of binarizing continuous representations to compact binary codes. A notable advantage of joint optimization is that we can improve the \emph{quantizability} of the learned deep representations $\{{\bm z}^*_i\}$ such that they can be quantized more effectively by our weak-orthogonal quantizer \eqref{eqn:Q}, yielding more accurate binary codes.

Approximate nearest neighbor (ANN) search by maximum inner-product similarity is a powerful tool for quantization methods \cite{cite:arxiv14CQIPS}. Given a database of $N$ binary codes $\{{\bm b}_n\}_{n=1}^{N}$, we follow \cite{cite:AAAI16DQN,cite:CVPR17DVSQ} to adopt \emph{Asymmetric Quantizer Distance} (AQD)  as the metric, which computes the inner-product similarity between a given query ${\bm q}$ and the reconstruction of the database point ${\bm x}_{n}$ as
\begin{equation}\label{eqn:AQD}
	\text{AQD}\left( {{{\bm q}},{{\bm x}_n}} \right) = {{\bm z}_q^{\mathsf T}} \left({\sum\limits_{m = 1}^M {{\bm C}_m{\bm b}_{mn}} }\right) ,
\end{equation}
Given query ${\bm q}$ and the deep representation ${\bm z}_q$, these inner-products between ${\bm z}_q$ and all $M$ codebooks $\{{\bm C}_m\}_{m=1}^M$ and all $K$ possible values of ${\bm b}_{mn}$ can be pre-computed and stored in a query-specific $M \times K$ lookup table, which is used to compute AQD between the query and all database points, each entails $M$ table lookups and additions and is slightly more costly than computing the Hamming distance.

\newcommand\mycommfont[1]{\footnotesize\ttfamily{#1}}
\SetCommentSty{mycommfont}
\begin{algorithm}[!tp]
    \DontPrintSemicolon
    \KwIn{$N$ training images $\mathcal{X} = \{{\bm x}_i\}_{i=1}^N$;}
    \KwIn{Similarity pairs $\mathcal{S} = \{{\bm s}_{ij}\}_{i,j=1}^N$.}
    \For{epoch $ = 0$ \KwTo \emph{MAX\_EPOCH}}{
    	Run the model to update $\{{\bm z}_i\}_{i=1}^{N}$ for $N$ training images\;
    	\If{epoch $ == 0$}{
    		Initialize $\bm{B}$ and $\bm{C}$ via Product Quantization \cite{cite:TPAMI11PQ}
    	}
    	Split the $N$ training images to $N/|G|$ groups randomly\;
    	$\mathcal{T} \leftarrow \emptyset$ \;
    	\For{\emph{group} $g = 0$ \KwTo $N/|G|$}{
    		\ForEach{$\bm{x}^a,\bm{x}^p \in G_g$, \textbf{s.t} $s_{ap} = 1$ }{
    			
    			\ForEach{$\bm{x}^n \in G_g$, \textbf{s.t} $s_{an} = 0$}{
    				
    				\If{$\delta  - \left\| {{\bm{z}}^a-{\bm{z}}^n} \right\|_2^2 + \left\| {{\bm{z}}^a-{\bm{z}}^p} \right\|_2^2 > 0$}{
    					\tcp{Triplet $<\bm{x}^a, \bm{x}^p, \bm{x}^n>$ is hard}
    					$\mathcal{T}_{ap} \leftarrow \mathcal{T}_{ap}~\cup \{<\bm{x}^a, \bm{x}^p, \bm{x}^n>\}$\;
    				}
    			}
    			\tcp{Randomly choose a hard negative sample from $\mathcal{T}_{ap}$}
    			$\mathcal{T} \leftarrow \mathcal{T} \cup \rm{rand}(\mathcal{T}_{ap})$\;
    		}
    	}
    	\For{i $ = 0$ \KwTo $|\mathcal{T}|/$\emph{BATCH\_SIZE}}{
    		Train the model using the $i$-th batch of triplets\;
    	}
    	Update $\bm{C}$ and $\bm{B}$ with Eqn. \eqref{eqn:updateC} and Eqn. \eqref{eqn:updateB}\ respectively\;
    	
    	\If{$|\mathcal{T}| < $ \emph{MIN\_TRIPLETS} {\bf and} $|G| > 1$}{
    		\tcp{Halve the size of the groups}
    		$|G| \leftarrow \lfloor \frac{|G|}{2} \rfloor$\;
    	}
    }
    \KwOut{The trained deep neural networks of DTQ.}
	\caption{Deep Triplet Quantization (DTQ) Training}
	\label{algorithm:DTQ}
\end{algorithm}

\subsection{Learning Algorithm}

The DTQ optimization problem in Equation~\eqref{eqn:model} consists of three sets of variables: deep convolutional neural network parameters ${\Theta}$, shared codebook ${\bm C} = [{\bm C}_1, \ldots, {\bm C}_M]$, and binary codes ${\bm B}^*$. We adopt an alternating optimization paradigm \cite{cite:SIGIR16CCQ} which iteratively updates one variable with the remaining variables fixed.

\textbf{Learning $\Theta$.}
The network parameters $\Theta$ can be efficiently optimized via standard back-propagation (BP) algorithm. We adopt the automatic differentiation techniques in TensorFlow.

\textbf{Learning $\bm{C}$.}
We update codebook ${\bm C}$ by rewriting Equation~\eqref{eqn:model} with ${\bm C}$ as the unknown variables in matrix formulation as follows,
\begin{equation}
	\mathop {\min }\limits_{\bm{C}}  {\sum\limits_{* \in \{ a, p , n \} } \left\| {{\bm{Z}}^* - {\bm{C}} {\bm{B}}^* } \right\|_F^2 + \gamma {\left\| {{\bm{C}}^{\sf T}{{\bm{C}}} - {\bm{I}}} \right\|_F^2}}.
\end{equation}
We adopt the gradient descent to update ${\bm C}$, ${\bm{C}} \leftarrow {\bm{C}} - \eta \frac{{\partial Q\left( {\bm{C}} \right)}}{{\partial {\bm{C}}}}$, and
\begin{equation}\label{eqn:updateC}
  \frac{{\partial Q\left( {\bm{C}} \right)}}{{\partial {\bm{C}}}} = 2\sum\limits_{ *  \in \{ a,p,n\} } {{\bm{C}}{{\bm{B}}^ * }{{\bm{B}}^{ * {\sf T}}}}  - 2\sum\limits_{ *  \in \{ a,p,n\} } {{{\bm{Z}}^ * }{{\bm{B}}^{ * {\sf T}}}}  + 2\gamma {\bm{C}}\left( {2{{\bm{C}}^{\sf T}}{\bm{C}} - {\bm{I}}} \right)
\end{equation}
where $\eta$ is a learning rate. We can further speed up computation by first solving ${\bm C}$ with $\gamma=0$, which leads to an analytic solution ${\bm{C}} = \left[ {\sum_{* \in \{ a, p , n \} } { {\bm{Z}}^* {\bm{B}}^{*\mathsf{T}} }} \right]{\left[ {\sum_{* \in \{ a, p , n \} } {{ {\bm{B}}^* {\bm{B}}^{*\mathsf{T}} }} } \right]^{ - 1}}$, then updating ${\bm C}$ with this solution as the starting point of gradient descent.

\textbf{Learning $\bm{B}$.}
As each ${\bm b}^*_i$ is independent on the rest of $\{{\bm b}^*_{i'}\}_{{i'} \ne i}$, the optimization for ${\bm B}^*$ can be decomposed to 3$N_t$ subproblems,
\begin{equation}\label{eqn:updateB}
  \begin{gathered}
    \mathop {\min }\limits_{{\bm{b}}_i^*} {\left\| {{\bm{z}}_i^* - \sum\limits_{m = 1}^M {{{\bm{C}}_m}{\bm{b}}_{mi}^*} } \right\|^2} \\
    {\text{s.t.}} \quad {\left\| {{\bm{b}}_{mi}^*}  \right\|_0} = 1, {\bm{b}}_{mi}^* \in {\left\{ {0,1} \right\}^K}. \\
  \end{gathered}
\end{equation}
This is essentially a high-order Markov Random Field (MRF) problem. As the MRF problem is generally NP-hard, we resort to the Iterated Conditional Modes (ICM) algorithm \cite{cite:ICML14CQ} that solves $M$ indicators $\{{\bm b}^*_{mi}\}_{m=1}^M$ alternatively. Specifically, given $\{{\bm b}^*_{{m'}i}\}_{{m'} \ne m}$ fixed, we update ${\bm b}^*_{mi}$ by exhaustively checking all the codewords in the codebook ${\bm C}_m$, finding the specific codeword with mimimal objective in \eqref{eqn:updateB}, and setting the corresponding entry of ${\bm b}^*_{mi}$ as $1$ and the rest as $0$. The ICM algorithm is guaranteed to converge to local minima, and can be terminated if maximum iteration is reached.
And the training procedure of DTQ is summarized in Algorithm \ref{algorithm:DTQ}.

\begin{table*}[!t]
    \centering
    \addtolength{\tabcolsep}{1pt}
    \caption{Mean Average Precision (MAP) Results for Different Number of Bits on the Three Benchmark Image Datasets}
    \label{table:MAP}
    \begin{tabular}{c|cccc|cccc|cccc}
        \Xhline{1.0pt}
        \multirow{2}{30pt}{\centering Method} & \multicolumn{4}{c|}{NUS-WIDE} & \multicolumn{4}{c|}{CIFAR-10} & \multicolumn{4}{c}{MS-COCO} \\
        \cline{2-13}
        & 8 bits & 16 bits  & 24 bits  & 32 bits  & 8 bits & 16 bits  & 24 bits  & 32 bits  & 8 bits & 16 bits  & 24 bits  & 32 bits \\
        \hline
ITQ-CCA &   0.526   &   0.575   &   0.572   &   0.594   &    0.315 &   0.354   &   0.371   &   0.414    &   0.501 &   0.566   &   0.563   &   0.562 \\
BRE     &   0.550   &   0.607   &   0.605   &   0.608   &    0.306 &   0.370   &   0.428   &   0.438    &   0.535 &   0.592   &   0.611   &   0.622 \\
KSH     &   0.618   &   0.651   &   0.672   &   0.682   &    0.489 &   0.524   &   0.534   &   0.558    &   0.492 &   0.521   &   0.533   &   0.534 \\
SDH     &   0.645   &   0.688   &   0.704   &   0.711   &    0.356 &   0.461   &   0.496   &   0.520    &   0.541 &   0.555   &   0.560   &   0.564 \\
        \hline
CNNH    &   0.586   &   0.609   &   0.628   &   0.635   &       0.461  &   0.476   &   0.476   &   0.472    &   0.505 &   0.564   &   0.569   &   0.574 \\
DNNH    &   0.638   &   0.652   &   0.667   &   0.687   &       0.525  &   0.559   &   0.566   &   0.558    &   0.551 &   0.593   &   0.601   &   0.603 \\
DHN     &   0.668   &   0.702   &   0.713   &   0.716   &    0.512 &   0.568   &   0.594   &   0.603    &   0.607 &   0.677   &   0.697   &   0.701 \\
HashNet &   0.613   &   0.662   &   0.687   &   0.699   &    0.621 &   0.643   &   0.660   &   0.667    &   0.625 &   0.687   &   0.699   &   0.718 \\
DQN     &   0.721   &   0.735   &   0.747   &   0.752 	&	0.527 &   0.551   &   0.558   &   0.564   &   0.649 &   0.653   &   0.666   &   0.685 \\
DVSQ    &    \underline{0.780}   &    \underline{0.790}   &    \underline{0.792}   &    \underline{0.797}   &    \underline{0.715}  &    \underline{0.727}   &    \underline{0.730}   &    \underline{0.733}    &   \underline{0.704}   &    \underline{0.712}   &    \underline{0.717}   &    \underline{0.720} \\
        \hline
DTQ     &  \textbf{0.795}   &   \textbf{0.798}   &   \textbf{0.799}   &   \textbf{0.801}    &    \textbf{0.785} &   \textbf{0.789}   &   \textbf{0.790}   &   \textbf{0.792}    &   \textbf{0.758}  &   \textbf{0.760}    &   \textbf{0.764}   &   \textbf{0.767} \\

        \Xhline{1.0pt}
    \end{tabular}
\end{table*}

\section{Experiments}

We conduct extensive experiments to evaluate the efficacy of the proposed DTQ approach against several state-of-the-art shallow and deep hashing methods on three image retrieval benchmark datasets, NUS-WIDE, CIFAR-10, and MS-COCO. Project codes and detailed configurations will be available at \url{https://github.com/thuml}.

\subsection{Setup}

The evaluation is conducted on three widely used image retrieval benchmark xdatasets: NUS-WIDE, CIFAR-10, and MS-COCO.

\textbf{NUS-WIDE}\footnote{\url{http://lms.comp.nus.edu.sg/research/NUS-WIDE.htm}} \cite{cite:CIVR09NusWide} is a public image dataset which contains 269,648 images in 81 ground truth categories.
We follow similar experimental protocols in \cite{cite:AAAI16DQN,cite:CVPR17DVSQ}, and randomly sample 5,000 images as query points, with the remaining images used as the database and randomly sample 10,000 images from the database for training.

\textbf{CIFAR-10}\footnote{\url{http://www.cs.toronto.edu/kriz/cifar.html}} is a public dataset with 60,000 tiny images in 10 classes. We follow the protocol in \cite{cite:AAAI16DQN} to randomly select 100 images per class as the query set, 500 images per class for training, and the rest images as the database.

\textbf{MS-COCO}\footnote{\url{http://mscoco.org}} \cite{cite:MSCOCO} is a dataset for image recognition, segmentation and captioning. The current release contains 82,783 training images and 40,504 validation images, where each image is labeled by some of the 80 semantic concepts. 
We randomly sample 5,000 images as the query points, with the rest used as the database, and randomly sample 10,000 images from the database for training.

Following standard evaluation protocol as previous work \cite{cite:AAAI14CNNH,cite:CVPR15DNNH,cite:AAAI16DHN,cite:CVPR17DVSQ,cite:ICCV17HashNet}, the similarity information for hash function learning and for ground-truth evaluation is constructed from image labels: if two images $i$ and $j$ share at least one label, they are similar and $s_{ij}=1$,  otherwise they are dissimilar and $s_{ij}=0$.
Though we use the ground truth image labels to construct the similarity information, the proposed DTQ can learn compact binary codes when only the similarity information is available, more general than label-based hashing and quantization methods \cite{cite:AAAI16DQN,cite:CVPR17DVSQ}.

We compare the retrieval performance of \textbf{DTQ} with ten state-of-the-art hashing methods, including supervised shallow hashing methods \textbf{BRE} \cite{cite:NIPS09BRE}, \textbf{ITQ-CCA} \cite{cite:CVPR11ITQ}, \textbf{KSH} \cite{cite:CVPR12KSH}, \textbf{SDH} \cite{cite:CVPR15SDH} and supervised deep hashing methods  \textbf{CNNH} \cite{cite:AAAI14CNNH}, \textbf{DNNH} \cite{cite:CVPR15DNNH}, \textbf{DHN} \cite{cite:AAAI16DHN}, \textbf{DQN} \cite{cite:AAAI16DQN}, \textbf{HashNet} \cite{cite:ICCV17HashNet}, \textbf{DVSQ} \cite{cite:CVPR17DVSQ}. 
We evaluate retrieval quality based on three standard evaluation metrics:  Mean Average Precision (\textbf{MAP}), Precision-Recall curves (\textbf{PR}), and Precision curves with respect to the numbers of top returned samples (\textbf{P@N}). To enable a direct comparison to the published results, all methods use identical training and test sets. We follow \cite{cite:AAAI16DQN,cite:ICCV17HashNet,cite:CVPR17DVSQ} and adopt MAP@5000 for NUS-WIDE dataset,  MAP@5000 for   MS-COCO dataset, and MAP@54000 for CIFAR-10 dataset.

Our implementation of DTQ is based on \textbf{TensorFlow}.
For shallow hashing methods, we use as image features the 4096-dimensional DeCAF$_7$ features \cite{cite:ICML14DeCAF}. 
For deep hashing methods, we use as input the original images, and adopt \emph{AlexNet} \cite{cite:NIPS12CNN} as the backbone architecture.
We fine-tune layers \textit{conv1}--\textit{fc7} copied from the AlexNet model pre-trained on ImageNet and train the last hash layer via back-propagation.
As the last layer is trained from scratch, we set its learning rate to be 10 times that of the lower layers. 
We use mini-batch stochastic gradient descent (SGD) with 0.9 momentum as the solver, and cross-validate the learning rate from $10^{-5}$ to $10^{-2}$ with a multiplicative step-size ${10}^{\frac{1}{2}}$. 
We fix $K = 256$ codewords for each codebook as \cite{cite:CVPR17DVSQ}.
For each point, the binary code for all $M$ codebooks requires 
	$B = M \, \log_2 \, K = 8M$ 
	bits (i.e. $M$ bytes), where we set $M = B/8$ as $B$ is a hyper-parameters.
We fix the mini-batch size of triplets as $128$ in each iteration and set the initial number of groups as $|G|=200$ for NUS-WIDE and MS-COCO, and $|G|=10$ for CIFAR-10.
We select the hyper-parameters of the proposed method DTQ and all comparison methods using the three-fold cross-validation.

\begin{figure*}[!thbp]
    \centering
    \subfigure[NUS-WIDE]{
        \includegraphics[width=0.275\textwidth]{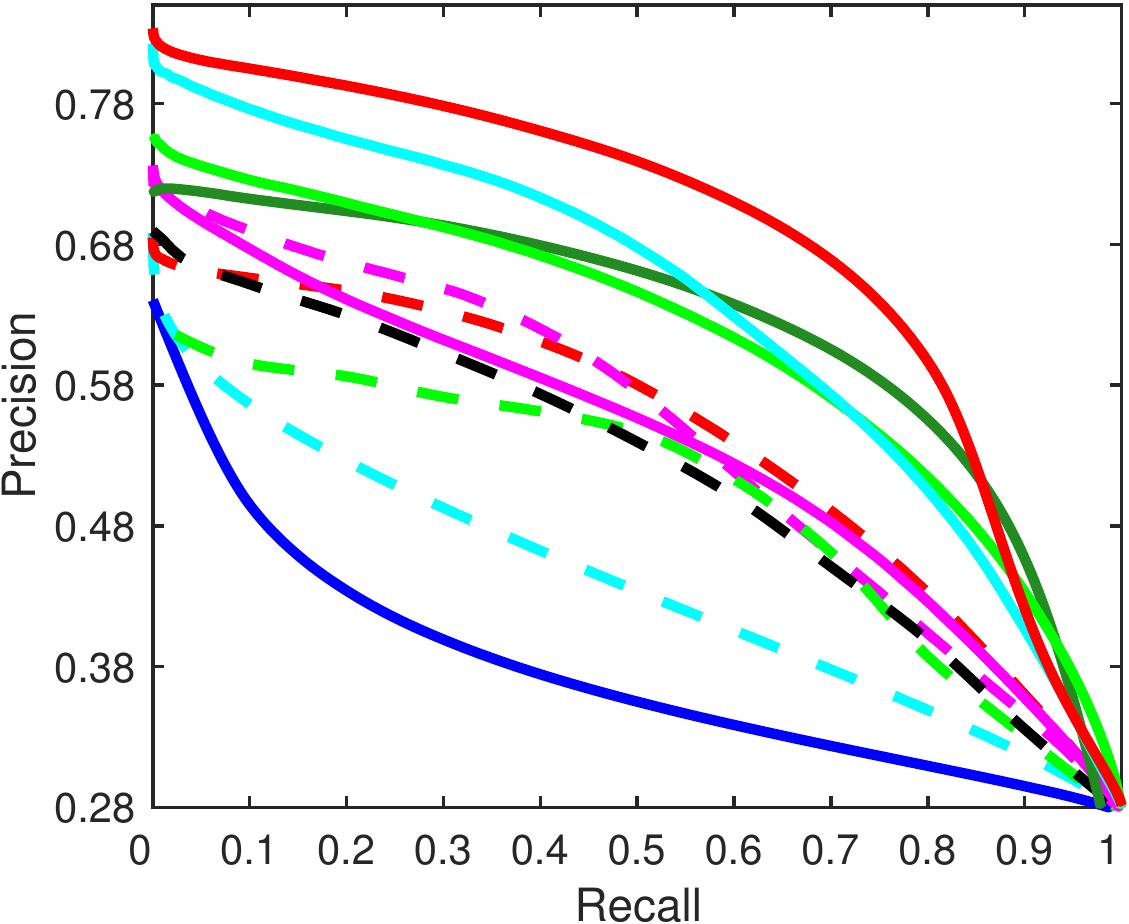}
        \label{fig:pr_nus}
    }
		\hfil
    \subfigure[CIFAR-10]{
        \includegraphics[width=0.275\textwidth]{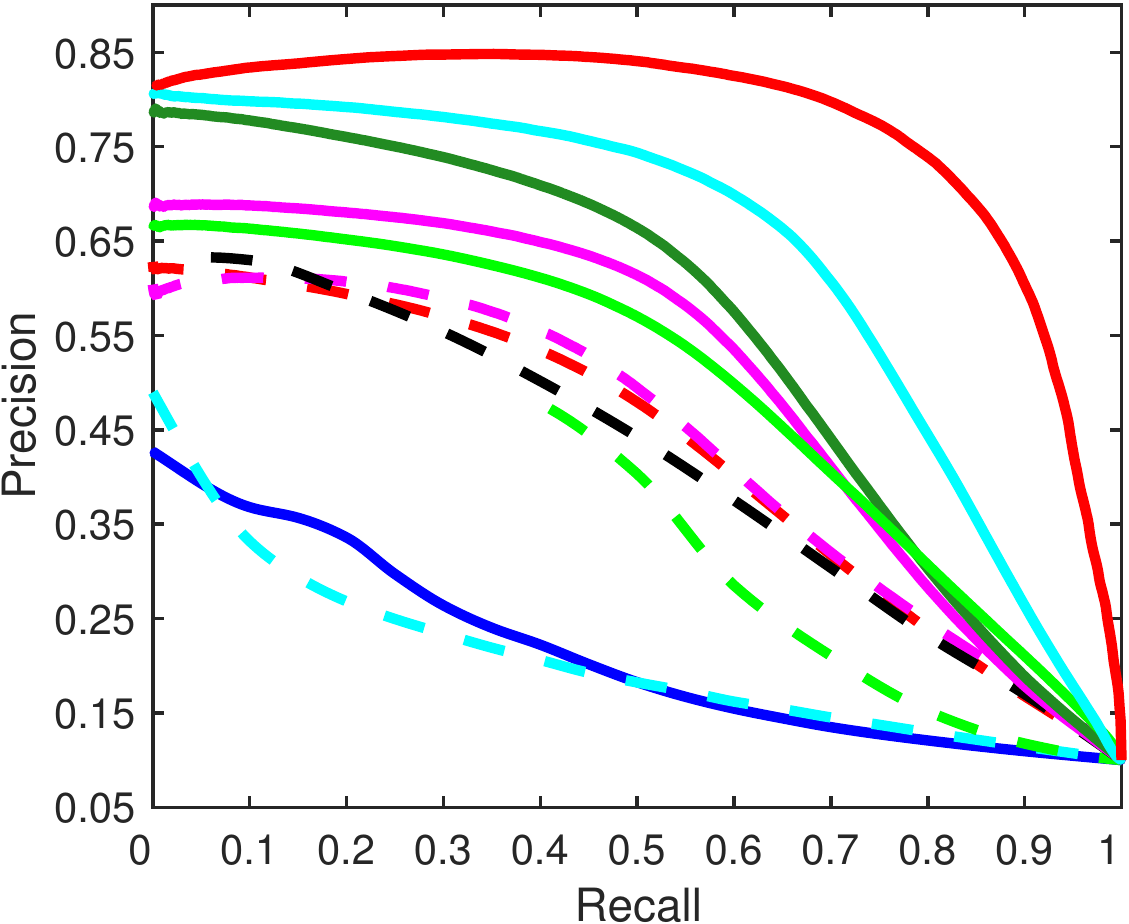}
        \label{fig:pr_cifar}
    }
		\hfil
	\subfigure[MS-COCO]{
        \includegraphics[width=0.36\textwidth]{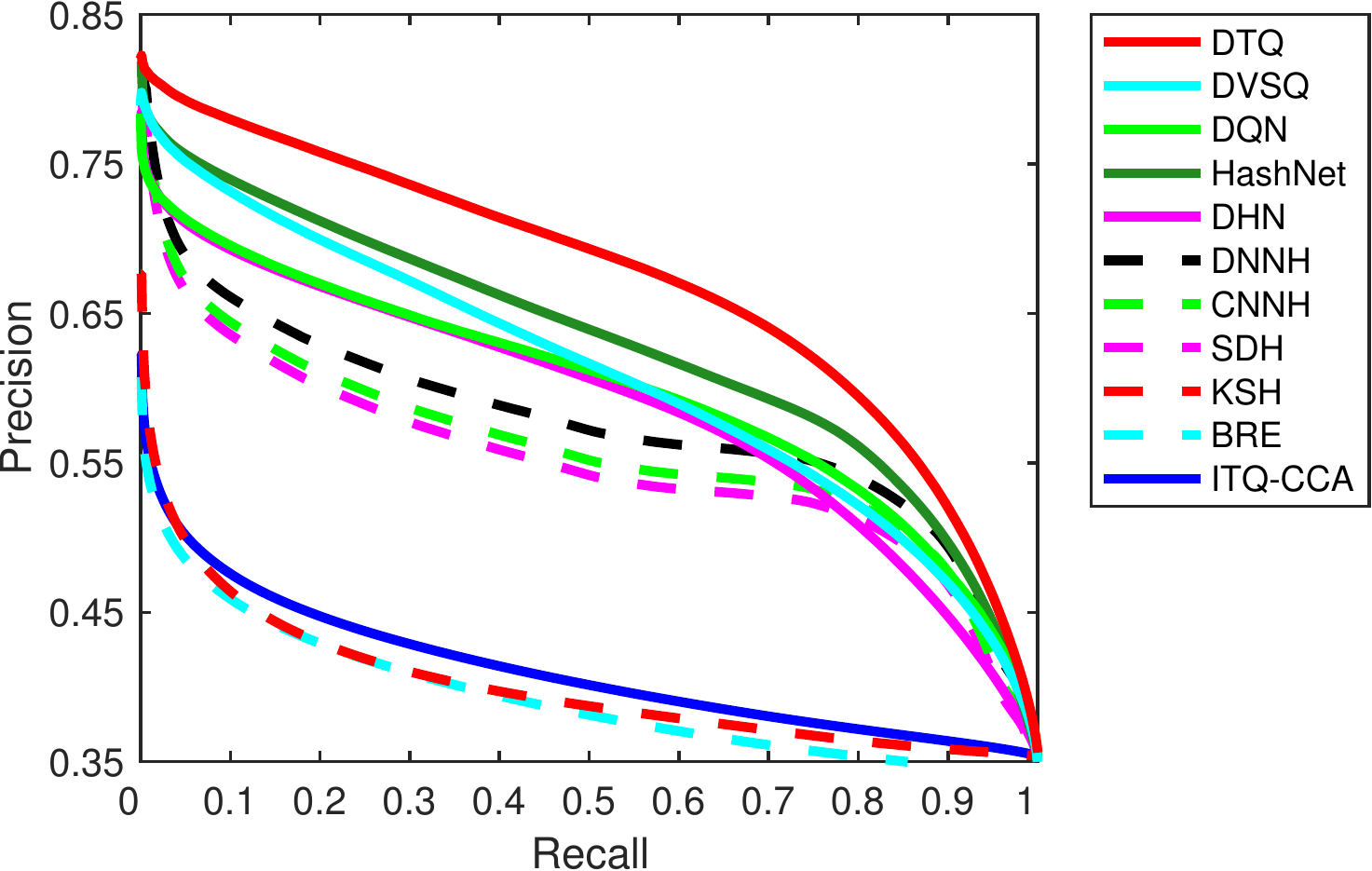}
        \label{fig:pr_coco}
    }
    \vspace{-5pt}
    \caption{Precision-recall curves on the NUS-WIDE, CIFAR-10 and MS-COCO datasets with binary codes @ 32 bits.}
    \label{fig:pr}
\end{figure*}

\begin{figure*}[!thbp]
    \centering
    \subfigure[NUS-WIDE]{
        \includegraphics[width=0.28\textwidth]{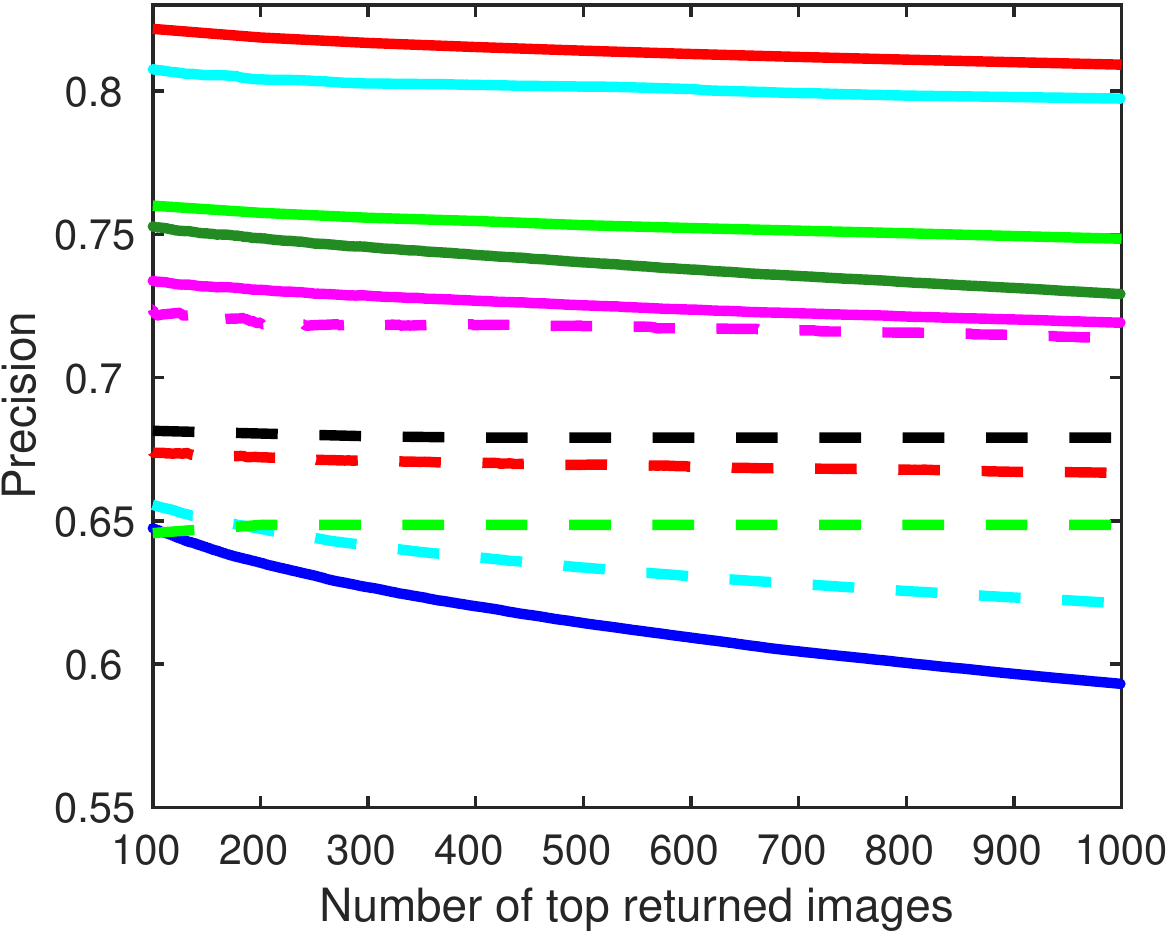}
        \label{fig:prec_nus}
    }
		\hfil
    \subfigure[CIFAR-10]{
        \includegraphics[width=0.28\textwidth]{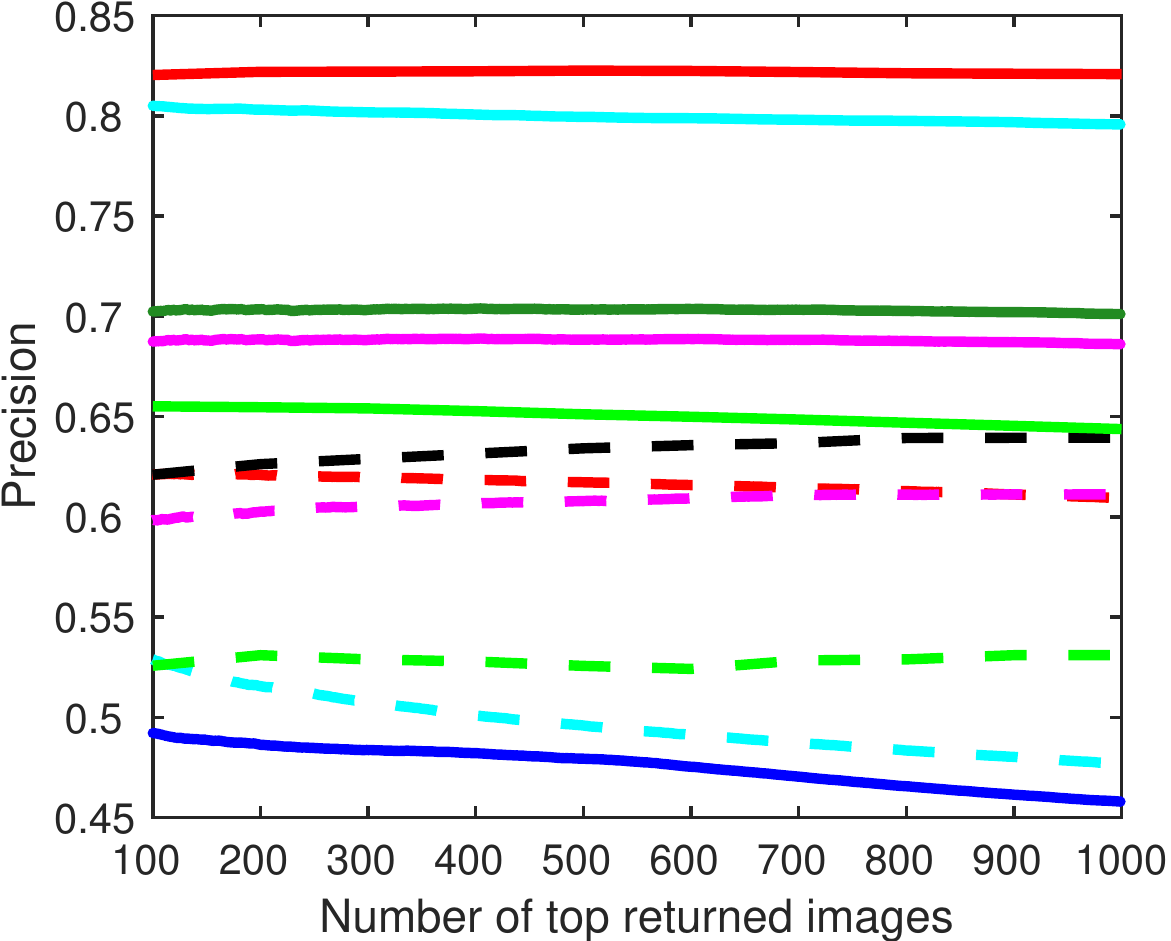}
        \label{fig:prec_cifar}
    }
		\hfil
    \subfigure[MS-COCO]{
        \includegraphics[width=0.36\textwidth]{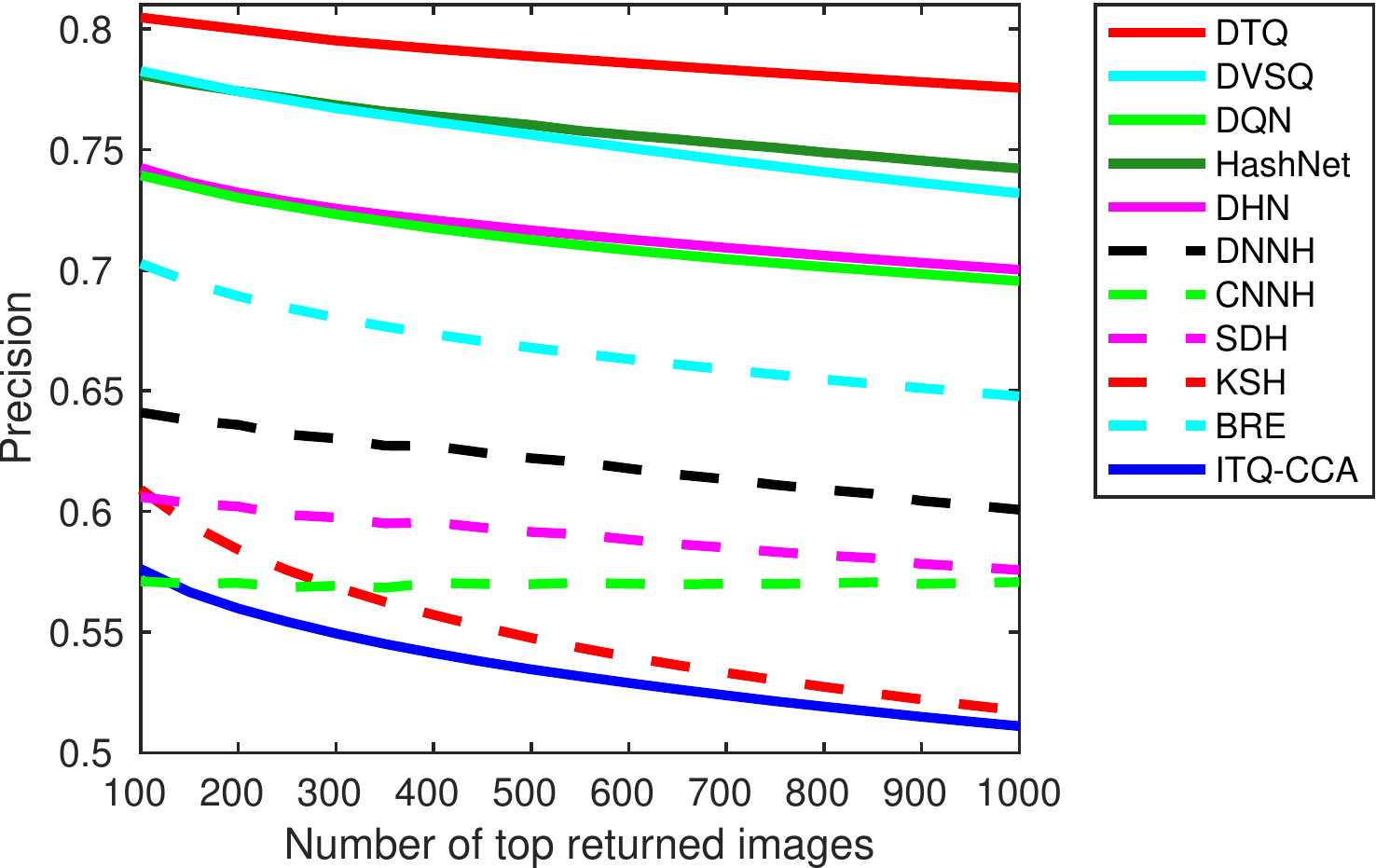}
        \label{fig:prec_coco}
    }
    \vspace{-5pt}
    \caption{Precision@top-N curves on the NUS-WIDE, CIFAR-10 and MS-COCO datasets with binary codes @ 32 bits.}
    \label{fig:prec}
\end{figure*}

\subsection{Results}

The \textbf{MAP} results of all methods are listed in Table \ref{table:MAP}, showing that the proposed DTQ substantially outperforms all the comparison methods. Specifically, compared to SDH \cite{cite:CVPR15SDH}, the best shallow hashing method with deep features as input, DTQ achieves absolute increases of \textbf{11.1\%}, \textbf{33.0\%} and {\textbf{20.7\%}} in the average MAP on NUS-WIDE, CIFAR-10, and MS-COCO respectively.
Compared to DVSQ \cite{cite:CVPR17DVSQ}, the state-of-the-art deep quantization method with class labels as supervised information, DTQ outperforms DVSQ by large margins of \textbf{0.8\%}, \textbf{6.2\%} and \textbf{4.9\%} in average MAP on the three datasets, NUS-WIDE, CIFAR-10, and MS-COCO, respectively. 

The MAP results reveal several interesting insights. \textbf{1)} Shallow hashing methods cannot learn discriminative deep representations and hash codes through end-to-end framework, which explains the fact that they are surpassed by deep hashing methods. \textbf{2)} Deep quantization methods DQN and DVSQ learn less lossy binary codes by jointly preserving similarity information and controlling the quantization error, significantly outperforming pioneering methods CNNH and DNNH without reducing the quantization error. 

The proposed DTQ improves substantially from the state-of-the-art DVSQ by three important perspectives: 
\textbf{1)} DTQ introduces a novel triplet training strategy to deep quantization framework for efficient similarity retrieval. It is worth noting that DTQ can learn compact binary codes when only the similarity information is available, which is more general than the label-based hashing method DVSQ.
\textbf{2)} During the learning of triplet loss, DTQ adopts a novel triplet mining strategy, \emph{Group Hard}, that mines appropriate amount of good triplets for each epoch, resulting in effective triplet training and better performance.
\textbf{3)} DTQ is the first method to apply weak-orthogonal quantization during triplet training. And back-propagating the triplet quantization loss can remarkably enhance the quantizability of the deep representations.

The retrieval performance in terms of Precision-Recall curves (PR) and Precision curves with respect to different numbers of top returned samples (\textbf{P@N}) are shown in Figures \ref{fig:pr} and \ref{fig:prec}, respectively. These metrics are widely used in deploying practical systems. The proposed DTQ significantly outperforms all the comparison methods by large margins under these two evaluation metrics. In particular, DTQ achieves much higher precision at lower recall levels or smaller number of top samples than all compared baselines. This is very desirable for precision-oriented retrieval, where people count more on the top-$N$ returned results with a small $N$. This justifies the value of our model for practical retrieval systems.

\begin{table*}[!tbh]
    \centering
    \addtolength{\tabcolsep}{1.0pt}
    \caption{Mean Average Precision (MAP) Results of DTQ and Its Variants DTQ-H, DTQ-T, DTQ-2, and DTQ}
    \label{table:ablation}
    \begin{tabular}{c|cccc|cccc|cccc}
        \Xhline{1.0pt}
        \multirow{2}{30pt}{\centering Method} & \multicolumn{4}{c|}{NUS-WIDE} & \multicolumn{4}{c|}{CIFAR-10} & \multicolumn{4}{c}{MS-COCO} \\
        \cline{2-13}
       	& 8 bits & 16 bits  & 24 bits  & 32 bits  & 8 bits & 16 bits  & 24 bits  & 32 bits  & 8 bits & 16 bits  & 24 bits  & 32 bits \\
        \hline
        DTQ-H   &   0.753  &    0.758  &    0.763 &    0.769  &      0.741    &   0.747  &    0.751  &    0.754      &      0.708    &    0.714  &    0.722  &    0.729   \\
        DTQ-T   &   0.719  &    0.722  &    0.727  &    0.731  &      0.663    &    0.670  &    0.672  &    0.679      &      0.714    &    0.720  &    0.728  &    0.734   \\
        DTQ-2   &   0.752  &    0.757  &    0.761  &    0.768  &      0.718    &    0.722  &    0.726  &    0.731      &     0.717    &    0.725  &    0.733  &    0.739   \\
        DTQ-Q   &   {0.769}  &    {0.773}  &    {0.777}  &    {0.781}  &      {0.750}    &    {0.761}  &    {0.763}  &    {0.765}      &      {0.721}    &    {0.727}  &    {0.734}  &    {0.740}   \\
        DTQ-O   &   \underline{0.785}  &    \underline{0.787}  &    \underline{0.780}  &    \underline{0.788}  &      \underline{0.771}    &    \underline{0.777}  &    \underline{0.779}  &    \underline{0.781}      &      \underline{0.739}    &    \underline{0.745}  &    \underline{0.750}  &    \underline{0.758}   \\
        DTQ     &   \textbf{0.795}  &    \textbf{0.798}  &    \textbf{0.799}  &    \textbf{0.801}  &      \textbf{0.785}    &    \textbf{0.789}  &    \textbf{0.790}  &    \textbf{0.792}      &      \textbf{0.758}    &    \textbf{0.760}   &    \textbf{0.764}  &    \textbf{0.767}   \\
        \Xhline{1.0pt}
    \end{tabular}
\end{table*}

\subsection{Analysis}

\subsubsection{Ablation Study}

We investigate five variants of DTQ:
\textbf{1) DTQ-T} is the DTQ variant by replacing the triplet loss in \eqref{eqn:L} with the widely-used pairwise cross-entropy loss \cite{cite:AAAI16DHN,cite:ICCV17HashNet};
\textbf{2) DTQ-H} is the DTQ variant without Group Hard to mine appropriate amount of good triplets for each epoch during the learning of the triplet loss as \cite{cite:CVPR15DNNH};
\textbf{3) DTQ-2} is the two-step variant of DTQ which first learns the deep representations for all images and then generates compact binary codes via the weak-orthogonal quantization.
\textbf{4) DTQ-Q} is the DTQ variant which replaces the proposed Triplet Quantization to the Product Quantization \cite{cite:TPAMI11PQ} used in DQN \cite{cite:AAAI16DQN}.
\textbf{5) DTQ-O} is the DTQ variant by removing the weak orthogonality penalty for redundancy reduction, i.e. $\gamma = 0$.

The MAP results for DTQ and it's five variants with respect to different code lengths on three benchmark datasets, NUS-WIDE, CIFAR-10, and MS-COCO are reported in Table \ref{table:ablation}.

\textbf{Triplet Loss.}
DTQ outperforms DTQ-T by very large margins of 7.4\%, 11.8\% and 3.8\% in the average MAP on the three datasets, NUS-WIDE, CIFAR-10, and MS-COCO, respectively.
DTQ-T uses the widely-used pairwise cross-entropy loss \cite{cite:AAAI16DHN,cite:ICCV17HashNet} which achieves state-of-the-art results on previous similarity retrieval tasks.
It is worth noting that the triplet loss is a learning to rank method, and tries to bring the anchor and the positive samples closer while also pushing away the negative samples. The DTQ with triplet loss is actually more suitable for the similarity retrieval tasks and naturally gives rise to much better performance than DTQ-T.

\textbf{Quantizability.} Another observation is that by jointly preserving similarity information in the deep representations of image triplets as well as controlling the quantization error of compact binary codes, DTQ outperforms DTQ-2 by 3.9\%, 6.4\% and 3.4\% in the average MAP on the three datasets, NUS-WIDE, CIFAR-10, and MS-COCO. This shows that end-to-end quantization can improve the quantizability of deep feature representations and satisfactorily yield much more accurate retrieval results.

\textbf{Triplet Quantization.} 
After replacing the proposed Triplet Quantization to Product Quantization \cite{cite:TPAMI11PQ} used in DQN \cite{cite:AAAI16DQN}, DTQ-Q yields significantly lossy compression and incur remarkable performance drop of 2.3\%, 2.9\%, 3.2\% in the average MAP on the three datasets, NUS-WIDE, CIFAR-10, and MS-COCO datasets respectively. This proves that the proposed Triplet Quantization with weak orthogonality can effectively learn compact binary codes and enable more effective retrieval than Product Quantization.

\textbf{Weak-Orthogonal Quantization.} 
Finally, by removing the weak orthogonality penalty, DTQ-O incurs performance drop of 1.3\%, 1.2\%, 1.4\% in the average MAP on the three datasets, NUS-WIDE, CIFAR-10, and MS-COCO datasets respectively. This proves the importance of removing the codebook redundancy and improving the compactness of binary codes for efficient image retrieval.

\begin{figure}[!thp]
    \centering
    \subfigure[Loss]{
        \includegraphics[width=0.48\columnwidth]{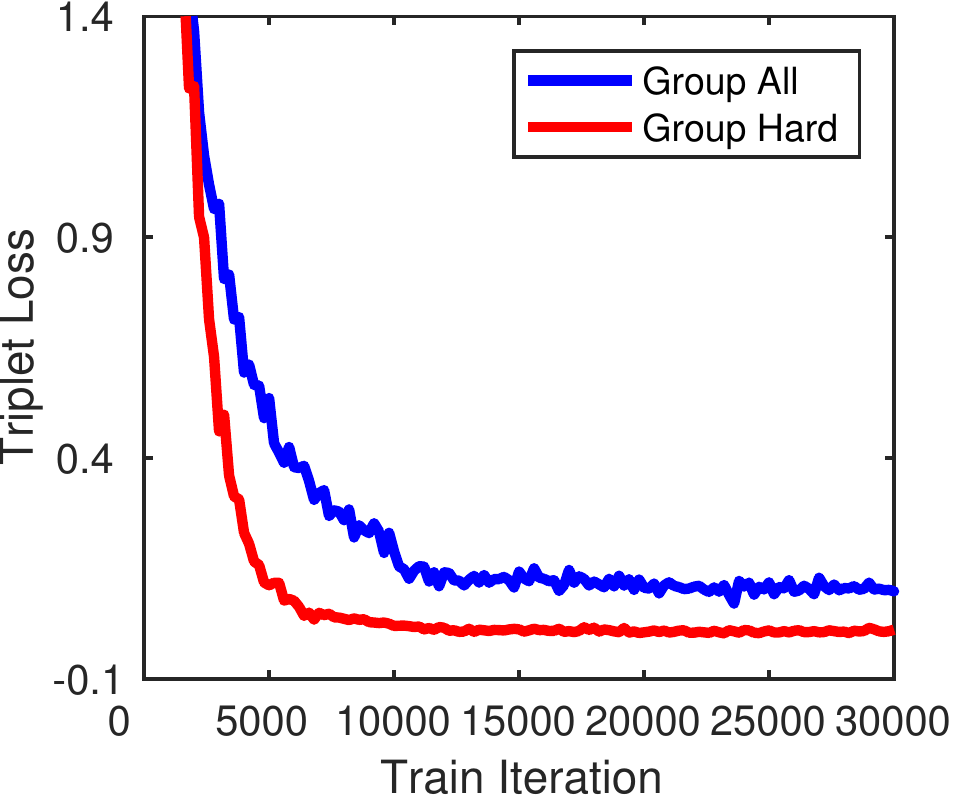}
        \label{fig:loss}
    }\hfil
    \subfigure[mAP]{
        \includegraphics[width=0.48\columnwidth]{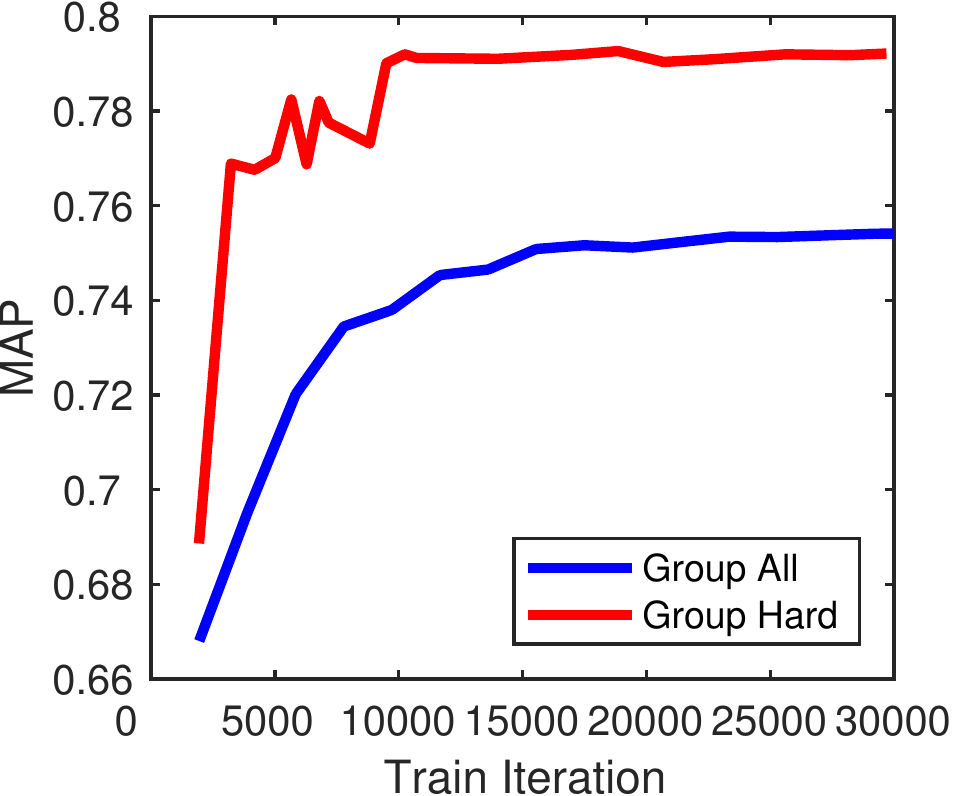}
        \label{fig:map}
    }
    \caption{Triplet Loss and MAP curves w.r.t. \#iterations.}
    \label{fig:loss_map}
\end{figure}

\subsubsection{Triplet Selection}

By using the proposed triplet mining strategy, \emph{Group Hard}, DTQ outperforms DTQ-H by large margins of 3.8\%, 4.0\% and 4.4\% in the average MAP on three benchmark datasets, NUS-WIDE, CIFAR-10, and MS-COCO, respectively. As shown in Figure~\ref{fig:loss_map}, without mining the appropriate amount of hard triplets, the \emph{Group All} training of triplet loss will quickly stagnate, leading to suboptimal convergence quality and MAP results. The proposed triplet mining strategy, \emph{Group Hard}, randomly samples proper amount of useful triplets with hard examples from several randomly partitioned group, resulting in effective training and faster convergence as well as more accurate retrieval performance.

\begin{table}[h]
    \addtolength{\tabcolsep}{2pt}
    \centering
    \caption{MAP on CIFAR-10 for Different Number of Bits}
    \label{table:online}
    \begin{tabular}{c|cccc}
        \Xhline{1.0pt}
        Method & 8 bits & 16 bits & 24 bits & 32 bits \\
        \hline
            DTQ-online & 0.703 & 0.708 & 0.710 & 0.713 \\
            DTQ & \bf{0.785} & \bf{0.789} & \bf{0.790} & \bf{0.792} \\
        \Xhline{1.0pt}
    \end{tabular}
\end{table}

\textbf{Online Selection.} Selecting all batch samples as negative is also known as \emph{online} triplet selection in the literature. Here we conduct a new experiment which uses online triplet selection and selects all hard negative samples in a batch (samples per batch = 192) for each anchor-positive pair. The results are reported in Table \ref{table:online}.
Due to the low ratio of the valid hard triplets in each batch for triplet training, DTQ-online (with online triplet selection) fails to achieve satisfactory retrieval results compared with the proposed DTQ.

\begin{figure*}[!tbh]
    \centering
    \subfigure[DVSQ]{
        \includegraphics[width=0.26\textwidth]{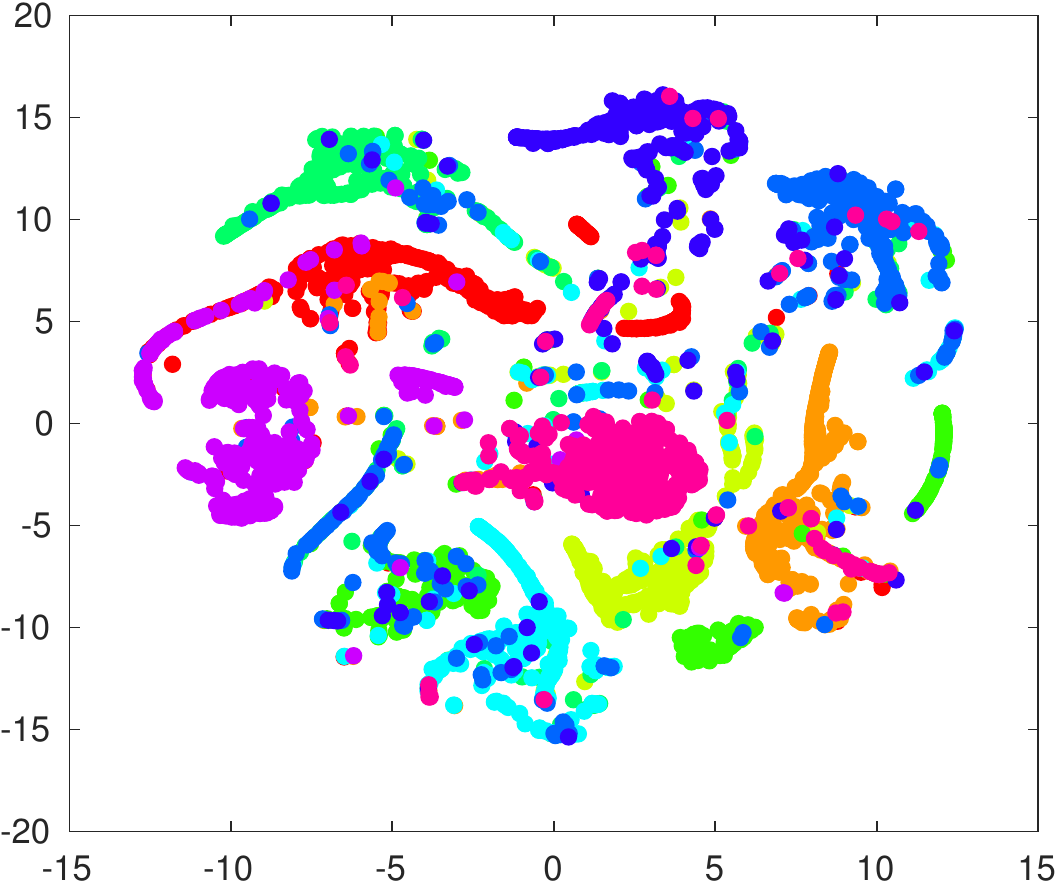}
        \label{fig:t-sne_dvsq}
    }\hfil
     \subfigure[DTQ-2]{
        \includegraphics[width=0.26\textwidth]{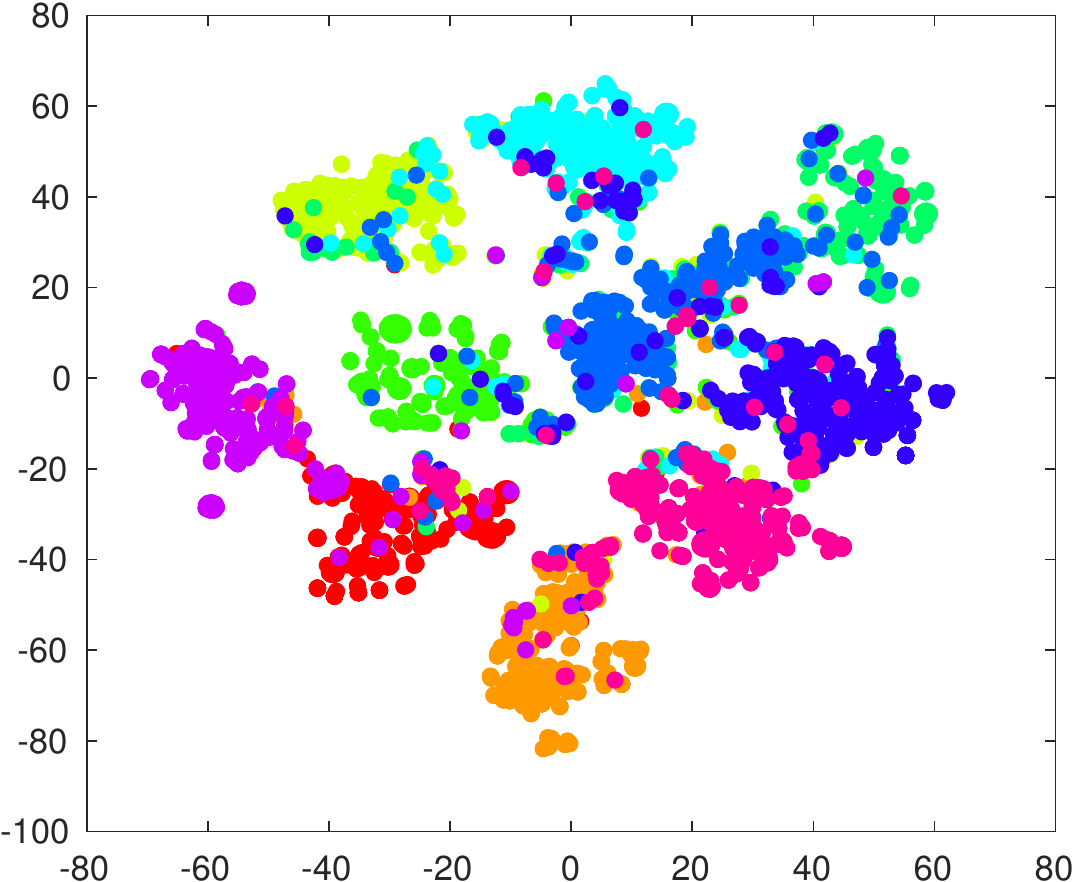}
        \label{fig:t-sne_dtq2}
    }\hfil
    \subfigure[DTQ]{
        \includegraphics[width=0.26\textwidth]{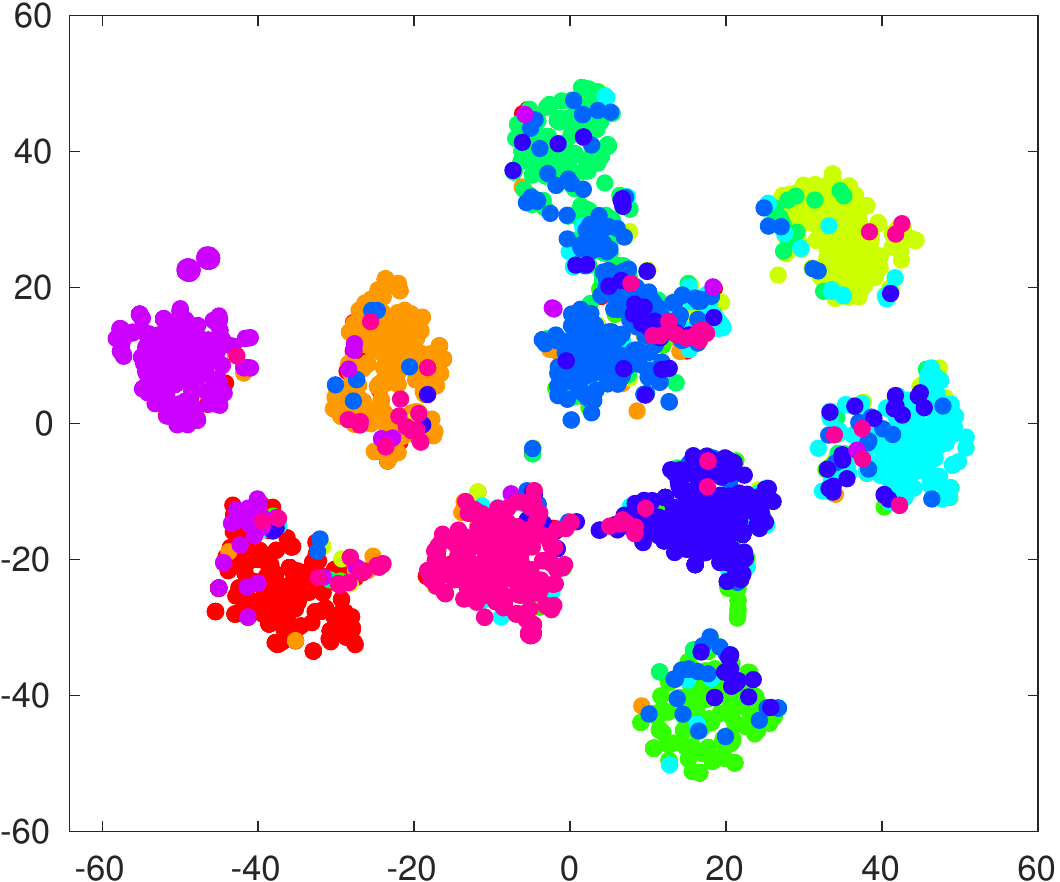}
        \label{fig:t-sne_dtq}
    }
    \vspace{-10pt}
	\caption{The t-SNE visualizations of deep representations learned by DVSQ, DTQ-2, and DTQ on CIFAR-10 dataset respectively.}
    \label{fig:tsne}
\end{figure*}

As online triplet selection cannot achieve satisfactory results, we adopt \emph{offline} triplet selection, which selects the valid hard triplets at the beginning of each epoch.
However, the offline strategy may generate too many candidate triplets and need a huge number of batches per epoch,
leading to hard triplets \emph{outdated} for training and potentially wasting most batches of each epoch.
To alleviate the outdated effect of hard triplets in offline selection, we split the data into specific groups and select hard triplets within each group, reducing the training triplets from $|\mathcal{T}|$ to $|\mathcal{T}|/|G|$. 

\begin{figure}[!tbh]
  \centering
  \includegraphics[width=0.46\columnwidth]{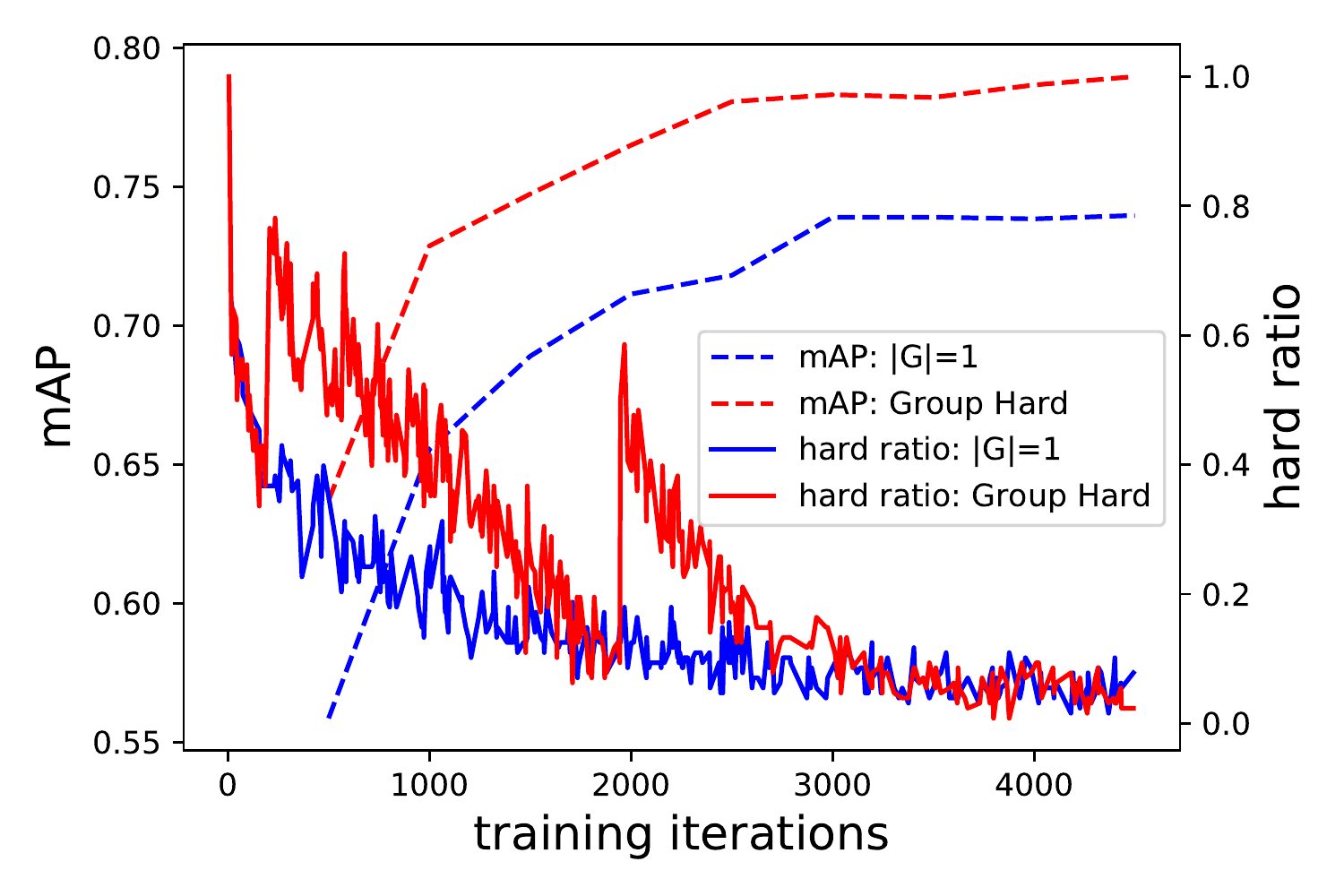}
  \includegraphics[width=0.44\columnwidth]{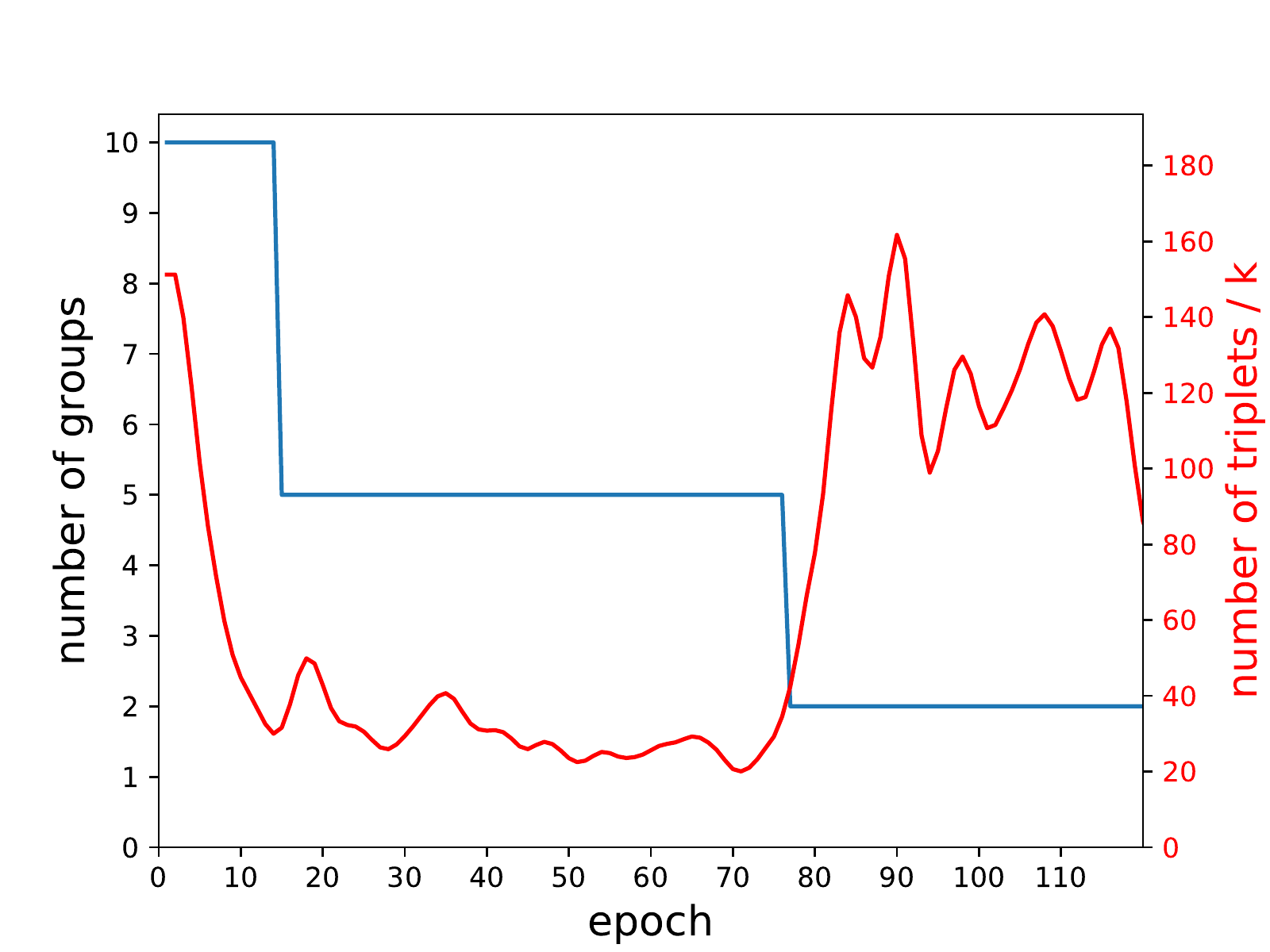}
  \caption{(left) mAP and ratio of non-outdated hard triplets w.r.t. iterations; (right) \#groups and \#triplets w.r.t. \#epochs.}
  \label{fig:n_part}
  \vspace{-10pt}
\end{figure}

We conduct an experiment to count the number of outdated hard triplets during training, shown in Figure~\ref{fig:n_part}. By splitting training data into $|G|$ specific groups, the number of outdated hard triplets is significantly reduced, leading to much better MAP results than the original offline triplet selection (i.e. $|G|=1$). This validates the effectiveness of the proposed offline selection strategy, \emph{Group Hard}.

\subsubsection{Visualization}

We show t-SNE visualization of binary codes and the illustration of top 10 returned images  for better understanding the impressive performance improvement of DTQ.

\textbf{Visualization of Representations.}
Figure \ref{fig:tsne} shows the t-SNE visualizations \cite{cite:tsne} of the deep representations learned by DVSQ \cite{cite:CVPR17DVSQ}, DTQ-2, and DTQ on CIFAR-10 dataset.
The deep representations of the proposed DTQ exhibit clear discriminative structures with data points in different categories well separated, while the deep representations by DVSQ \cite{cite:CVPR17DVSQ} exhibit relative vague structures.
This validates that by introducing the triplet training to deep quantization, the deep representations generated by our DTQ are more discriminative than that generated by DVSQ, enabling more accurate image retrieval.
Also, the deep representations of DTQ are more discriminative than that of the two-step variant DTQ-2, showing the efficacy of jointly preserving similarity information in the deep representations of image triplets and controlling the quantization error of compact binary codes via back-propagation.

\begin{figure}[!tbh]
  \centering
  \includegraphics[width=1.0\columnwidth]{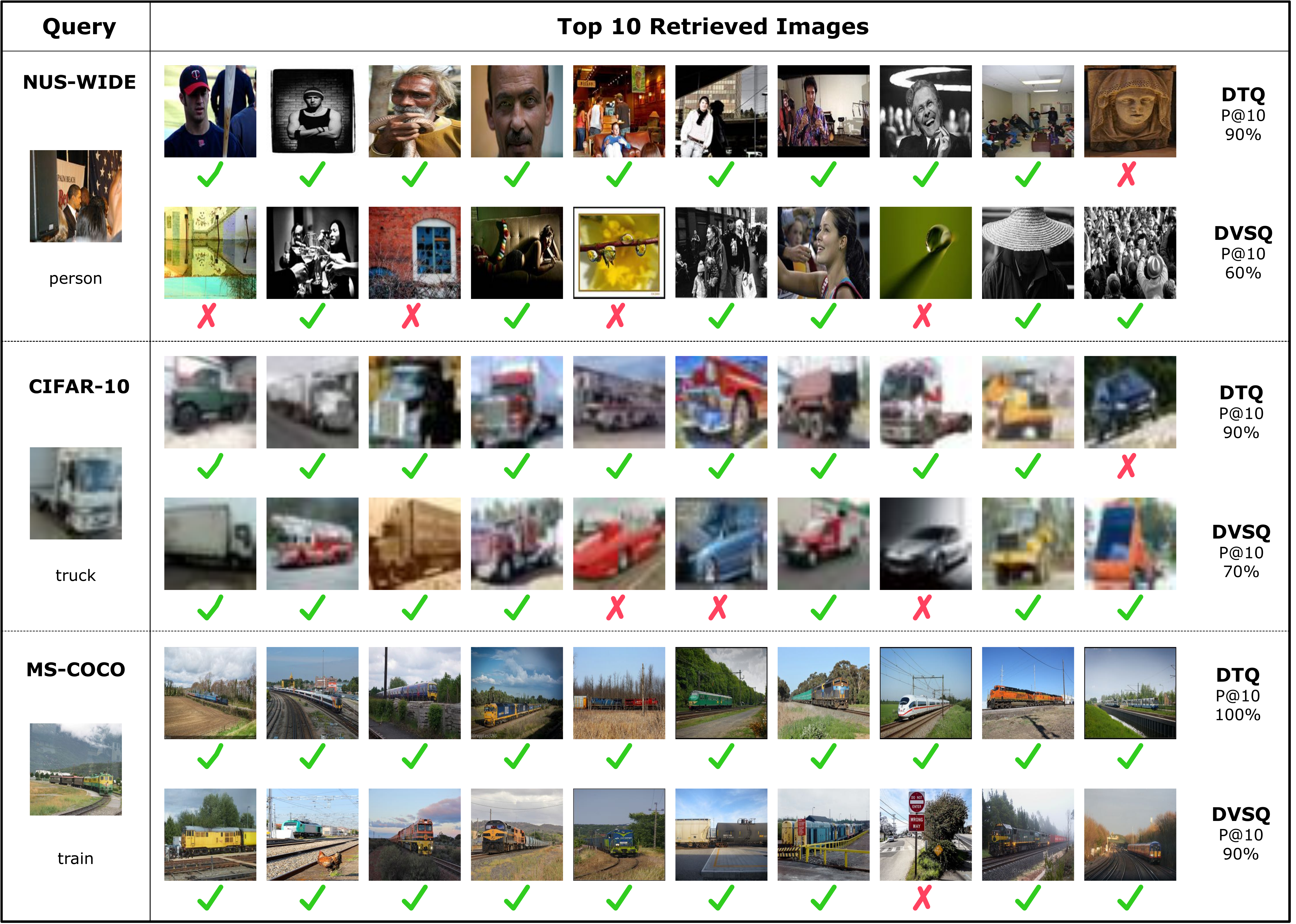}
  \caption{The top 10 images returned by DVSQ and DTQ.}
  \label{fig:top10}
  \vspace{-10pt}
\end{figure}

\textbf{Illustration of Top 10 Results.}
Figure \ref{fig:top10} illustrates the top 10 returned images of DTQ and the best deep hashing baseline DVSQ \cite{cite:CVPR17DVSQ} for three query images on the three datasets NUS-WIDE, CIFAR-10, and MS-COCO, respectively. DTQ yields much more relevant and user-desired retrieval results than the state-of-the-art method.

\section{Conclusion}

This paper proposed Deep Triplet Quantization (DTQ) for efficient image retrieval, which introduces a triplet training strategy to deep quantization framework. Through a novel triplet selection module, {Group Hard}, an appropriate number of hard triplets are selected for effective triplet training and faster convergence. To enable efficient image retrieval, DTQ can learn compact binary codes by jointly optimizing a novel triplet quantization loss with weak orthogonality. Comprehensive experiments justify that DTQ generates compact binary encoding and yields state-of-the-art retrieval performance on three benchmark datasets NUS-WIDE, CIFAR-10, and MS-COCO.

\section{Acknowledgements}

This work is supported by National Key R\&D Program of China (2016YFB1000701), and NSFC grants (61772299, 61672313, 71690231).

{
\bibliographystyle{ACM-Reference-Format}
\bibliography{DTQ}
}

\end{document}